\documentclass[letterpaper]{article} 
\usepackage{aaai2026}  
\usepackage{times}  
\usepackage{helvet}  
\usepackage{courier}  
\usepackage[hyphens]{url}  
\usepackage{graphicx} 
\usepackage{natbib}  
\usepackage{caption} 
\usepackage[x11names]{xcolor}
\usepackage{algorithm}
\usepackage{algorithmic}
\usepackage{arydshln}
\usepackage{tcolorbox}
\usepackage{subcaption}

\usepackage{colortbl}

\usepackage{mdframed}

\usepackage{pifont}
\newcommand{\cmark}{\ding{51}}
\newcommand{\xmark}{\ding{55}}
\newcommand*{\xmltag}[1]{\texttt{<#1>}}

\usepackage{newfloat}
\usepackage{listings}

\DeclareCaptionStyle{ruled}{labelfont=normalfont,labelsep=colon,strut=off} 
\floatstyle{ruled}
\newfloat{listing}{tb}{lst}{}
\floatname{listing}{Listing}
\title{Less is More: Learning Graph Tasks with Just LLMs}
\author{
    Sola Shirai,
    Kavitha Srinivas,
    Julian Dolby,
    Michael Katz,
    Horst Samulowitz,
    Shirin Sohrabi
}
\affiliations{
    IBM Research\\
    New York, USA\\
    \{solashirai, kavitha.srinivas, michael.katz1\}@ibm.com\\
    \{dolby, samulowitz, ssohrab\}@us.ibm.com
}

\nocopyright

\begin{document}

\maketitle

\newcommand{\inlinecite}[1]{\citeauthor{#1}~\shortcite{#1}}

\begin{abstract}

For large language models (LLMs), reasoning over graphs could help solve many problems.
Prior work has tried to improve LLM graph reasoning by examining how best to serialize graphs as text and by combining GNNs and LLMs.
However, the merits of such approaches remain unclear, so we empirically answer the following research questions: (1) Can LLMs learn to solve fundamental graph tasks without specialized graph encoding models?, (2) Can LLMs generalize learned solutions to unseen graph structures or tasks?, and (3) What are the merits of competing approaches to learn graph tasks? We show that even small LLMs can learn to solve graph tasks by training them with instructive chain-of-thought solutions, and this training generalizes, without specialized graph encoders, to new tasks and graph structures. 
\end{abstract}


\section{Introduction}

For large language models (LLMs), reasoning over graphs helps solve many problems.  For instance, answering questions that combine graphs with textual data -- in knowledge graph question answering, e.g. KGQA\cite{ji-etal-2024-retrieval}, graph-RAG \cite{Edge2024FromLT}, or code localization and search \cite{chen2025locagentgraphguidedllmagents}.   Because this ability is often limited in pre-trained LLMs, approaches often serialize the neighborhood structure of a node into natural language so that it can be unified with textual data.  This usually works well for smaller neighborhoods (1 or 2 hops) but often breaks down for larger ones.  Hence it would be ideal if LLMs could process reasonable sized graphs directly.

There two approaches to adding graph knowledge explicitly to an LLM\footnote{We exclude approaches that try to generalize a graph neural network to reason across multiple tasks here, because that is not a case where a pretrained LLM is leveraged to answer questions over graph and textual data}: 1. Encode graphs with a specialized encoder, and fuse those embeddings into the LLM, e.g. \cite{chai2023graphllmboostinggraphreasoning}; 2. With instruction tuning, train LLMs to reason over graphs serialized as text, e.g., \cite{10.1145/3637528.3672010}.

Previous comparisons of these approaches have been limited; further, the graphs used for training and testing have often been small (20 or fewer nodes), and the generalizability to new graph tasks and graph sizes remains unclear.  So we address the following research questions: 
\begin{itemize}
    \item Can LLMs learn to solve fundamental graph tasks without specialized graph encoders?
    \item Can LLMs generalize from these learned solutions to graph structures or tasks that are unseen at training?
    \item What are the tradeoffs and limitations of competing approaches to graph tasks? 
\end{itemize}

We compare 4 methods to infuse graph knowledge into LLMs: (a) \textbf{Graph Tokens}, which encode the graph with a graph transformer, and a projection layer to integrate with the LLM \cite{perozzi2024let}.  Training is on the transformer and projection layer; LLM weights are frozen. (b) \textbf{Graph Tokens + Text} where a textual representation of the graph is added, so training can establish their correspondence. (c) \textbf{LoRA}, which fine tunes the LLM \cite{DBLP:journals/corr/abs-2106-09685}, (d) \textbf{P-Tuning} as a soft prompt technique \cite{liu-etal-2022-p}, where the LLM gets the graph solely as text, and a continuous prompt encoding is learned during training. (a) and (b) infuse LLMs with graph knowledge, (c) fine tunes the LLM with textual forms of graphs, and (d) is a control for the graph tokens approaches because the LLM is given ``soft prompts'' to represent any extra information about the graph.

We train a small LLM model
\cite{microsoft2025phi4minitechnicalreportcompact} with four tasks that target graph structure: node count, node degree, and reachability using BFS or DFS. Our work suggests that LLMs can learn about graph structure and graph reachability without additional graph encoders. The tuned model generalizes to graph structures unseen during training, and generalizes to much larger graph sizes and paths, maintaining 88\% accuracy on graphs which are twice as large as those seen during training with 100-step DFS traces (75\% of training samples have 19 steps or fewer).

For out of distribution tasks related to reachability (e.g., cycle check or shortest path), trained models did not consistently show transfer; however, with just 30 examples of additional training, they showed a clear improvement -- suggesting some generalizability.  Infusing graph knowledge into LLMs helps it answer questions on knowledge graphs (e.g. in LC-QUAD~\cite{dubey2019lc}). More interestingly, the base LLM can generate graphs from a textual problem (e.g. in PrOntoQA~\cite{PrOntoQA}), which the tuned models use to show much better performance than answering the question from text alone.

\begin{figure*}[!h]
    \centering
    \includegraphics[width=.9\linewidth]{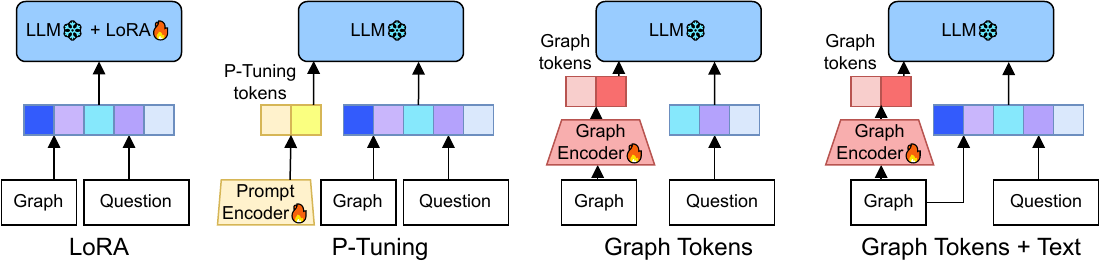}
    \caption{Overview of our 4 experimental models. The LLM's weights are left frozen for all approaches.}
    \label{fig:graph_llm_models}
\end{figure*}

\section{Related work}

\subsubsection{Limitations of LLMs for Graph Tasks}
Several papers \cite{liu2023evaluating,dai2024revisiting} find that even state of the art models struggle to solve simple graph tasks. \inlinecite{sanford2024understanding} and \inlinecite{abbe-et-al-neurips2024} describe theoretical limitations of learning graph tasks using transformer models, suggesting the difficulty of learning certain tasks and an inability to generalize to out-of-distribution graphs.
Our results show that standard supervised fine-tuning, together with instruction tuning and multi-task learning, can alleviate the generalization issue. 

Benchmark datasets have also been developed for better evaluation of this problem, such as NLGraph \cite{wang2023can}, GraphOmni \cite{xu2025graphomnicomprehensiveextendablebenchmark}, GraphQA \cite{sanford2024understanding}, and NLGift \cite{zhang-etal-2024-llm-graph}, but most have graphs with fewer than 50 nodes; our work explores training models on graphs with up to 100 nodes and further evaluate how well LLMs learn to solve tasks on out-of-distribution graphs and tasks. GraphWiz \cite{10.1145/3637528.3672010} introduces a new dataset for training on graphs called GraphInstruct, a training dataset which is similar to the GraphQA dataset but enhanced with instructions, as we have in our work. A key difference of our approach is that we produce instructions deterministically, while GraphWiz generates reasoning text using an LLM. Our work also differs in that we more carefully control the distribution of task answers.

\subsubsection{LLMs as Graph Reasoners}
LLaGa \cite{10.5555/3692070.3692376} and GraphEdit \cite{guo2025grapheditlargelanguagemodels} convert graphs into node sequences and answers questions such as node classification and link prediction; our work is geared more to understand how well LLMs understand graphs as a basic structure. 
PromptGFM \cite{zhu2025llmgnngraphvocabulary} similarly simulates a GNN workflow within an LLM, but as with LLaGa, the focus is not in testing graph reasoning directly in LLMs.

\subsubsection{Multi Model Approaches}
In Graph Tokens \cite{perozzi2024let} (see also \cite{chai2023graphllmboostinggraphreasoning}), GNNs are used to encode graphs into node or edge vectors, after which aggregation is used to produce a final set of embeddings to input to the LLM. Both of these works only explore very small graphs (less than 20 nodes). In addition, Graph Tokens uses a different GNN and tokenization architecture for each task. Performance of a random baseline is not clear in either work, making relative improvement unclear; and generalization is not measured systematically, as it is in our work. 


GraphGPT \cite{10.1145/3626772.3657775} uses a pretrained graph encoder and an alignment mechanism to project graph tokens to an LLM. The LLM is trained in an unsupervised manner to learn to output the order of node text to learn the alignment between text and the graph. GraphGPT was tested with knowledge graph completion tasks; the focus is not on testing graph reasoning in LLMs directly. GOFA \cite{kong2024gofa} creates a graph model integrated with LLMs by interweaving LLM compressors with GNN layers that capture the structure of the graph, targeting knowledge graph completion. 


\section{Method}


We trained four models, on four graph tasks.

\subsection{Models}

Each experimental model augments the base LLM without fully retraining its weights -- Figure \ref{fig:graph_llm_models} compares each model and how it interacts with the LLM. We use standard cross entropy loss as the loss function to perform causal language model training.

\subsubsection{LoRA} 

LoRA \cite{hu2022lora} is a fine-tuning approach for LLMs, where low-rank matrices are trained and inserted into layers of the LLM to adapt its weights to new tasks. 



\subsubsection{P-Tuning} P-Tuning \cite{liu2024ptuning} inserts a fixed number of learnable embeddings into the LLM together with the input prompt. Tokens produced by P-Tuning augment the prompt, providing embeddings as a ``prompt'' rather than discrete text. 




\subsubsection{Graph Tokens} Graph Tokens \cite{perozzi2024let} encode an input graph into a set of tokens with a GNN, which augment the question input to an LLM.  The GNN is trained based on the loss function of the LLM.  Unlike P-Tuning's static tokens, graph tokens differ for each unique input graph. 

\subsubsection{Graph Tokens + Text} This method combines Graph Tokens with textual descriptions of the graph (i.e., the same input as LoRA and P-tuning). This model type allows graph tokens created by the GNN to benefit from alignment with the textual representation of the graph.  

\subsection{Graph Question Answering Tasks}

We used 8 graph tasks inspired by the GraphQA benchmark \cite{sanford2024understanding}.

The easiest 4 are retrieval tasks, solvable by a simple lookup or aggregation: \textbf{Node Count} (NC) asks the number of nodes in the graph. \textbf{Node Degree} (ND) is the outgoing degree of a particular node. \textbf{Edge Existence} (EE) asks if an edge from node A to node B exists. \textbf{Edge Count} (EC) is the number of edges in the graph.

We investigated three perspectives on the reachability (or connectivity) task, which asks whether a path exists in the graph from node A to node B. We investigate \textit{how to solve} the task, using \textbf{Depth First Search} (DFS) or \textbf{Breadth First Search} (BFS). Additionally, \textbf{Cycle Check} (CC) asks if the graph has a cycle starting from a target node (i.e., a reachability task from node A to itself). Note that this differs slightly from related work, which typically ask if a graph contains \textit{any} cycle.

Our last task is \textbf{Shortest Path} (SP), which returns a list of nodes visited in the shortest path connecting node A to node B, or an empty list if there is none. We determine a single correct shortest path based on a BFS traversal, with nodes ordered by their node ID during traversal.

We used only four tasks at training (\textbf{NC}, \textbf{ND}, \textbf{DFS}, \textbf{BFS}), and the rest are used for testing to see if the model can generalize to new, but related, tasks. We also controlled the distribution of answers for tasks, to reduce class imbalance and control for the performance of a random chance baseline. 

Each task is asked in natural language, for which we use templates to produce several variations of each question. (Examples in Appendix 1).

\subsection{Textual Graph Representation}

\inlinecite{fatemi2024talk} show that the textual representation of a graph affects LLM performance. We investigate the performance of models across multiple tasks on a single representation style, with a mixture of insights from prior works.

Figure \ref{fig:graph_text_example} shows an example of our graph representation. Ellipses indicate prompt text omitted for brevity. First, nodes are indicated using integer IDs, and edges as an incidence list; this helps reduce context length. Second, nodes are associated with semantics using a dictionary mapping node IDs to labels. These labels are randomly generated words from an English dictionary, and serve to add some textual components to the graph while avoiding any implicit reliance on semantics associated with real text graphs. Last, in our question text, we use the node labels in the question rather than node IDs. 

\begin{figure}[h]
\centering
\begin{mdframed}[backgroundcolor=Ivory1]
\scriptsize
...

G describes a graph among nodes with the following mapping from IDs to labels:

\{0: visible,
1: askance,
2: oikoplast, ..., 5: sleepingly\} 

The edges in G are: 

0 $\to$ 1, 3

2 $\to$ 0, 4

...

Q: Is there a breadth first traversal path from ``oikoplast" to ``sleepingly"?
\end{mdframed}
\caption{Example representation of a graph and question.}
\label{fig:graph_text_example}
\end{figure}

We use this format to encode graphs for the input prompts for all models except Graph Tokens; there, we omit the text describing the graph's edges but include the ID to label mapping. While prior work has investigated how to encode semantic information about nodes into the graph encoding, in our preliminary studies we found that simple semantics such as random labels were not effectively encoded by the GNN. We therefore opted to use LLM prompting to add this semantic information about nodes instead\footnote{Several preliminary experiments showed poor performance on the graph tokens approach without this text.}.

\subsection{Reasoning Steps for Task Solutions}


Having an LLM produce a chain of thought (CoT) \cite{wei2022chain}, especially as a type of reasoning chain, has become a mainstay approach in enhancing the capabilities of LLMs. Our own experimental validation (see Appendix 2) demonstrated the utility of this approach for graph reasoning, so it was always included in training.

For each task, we implemented a template to produce reasoning steps deterministically from code as part of the training and validation data. At test time, accuracy is only judged based on the final answer. As in prior work, the LLM was instructed to output steps within \xmltag{think} tags and the final answer in \xmltag{answer} tags. 

Following \inlinecite{aytes2025sketch}, showing the effectiveness of concise sketches as CoT, our reasoning steps use a mixture of python-like lists and arrows to indicate elements such as edges and paths, along with natural language describing the approach at the beginning of the CoT. An example for our BFS task is shown in Figure \ref{fig:bfscot_text_example} -- ellipses indicate text omitted for brevity. A full listing of CoT text examples for our tasks can be found in Appendix 1.

\begin{figure}[h]
\centering
\begin{mdframed}[backgroundcolor=Ivory1]
\scriptsize
\xmltag{think} Starting breadth first traversal from oikoplast to see if sleepingly is reachable. ...

Queue: [2]

2 $\to$ 0, 4

Queue: [0, 4]

...

Queue: []

No unvisited nodes remain.\xmltag{think}

\xmltag{answer}No\xmltag{/answer}
\end{mdframed}
\caption{Instructive CoT for the BFS task.}
\label{fig:bfscot_text_example}
\end{figure}

\section{Experimental Setup}


\subsection{LLM Choice}

Our main experiments are performed using Phi4-Mini-Instruct \cite{microsoft2025phi4minitechnicalreportcompact}, a 4-billion parameter LLM. Phi4-Mini was selected due to strong performance shown during early experimentation, as well as suitability in terms of memory requirements versus performance. All training and inference performed with this model can fit within the memory limits of a single 80GB GPU.




\subsection{Datasets}

As in GraphQA \cite{sanford2024understanding}, we synthetically generate graph data and associated questions. All graph data used in our experiments are directed and unweighted. In order to address our research questions surrounding how well LLMs can learn to solve graph tasks as well as different facets of generalizability, our test datasets are constructed primarily with graph structures, graph sizes, and tasks that were unseen or poorly represented during training.

Unlike prior work, answer classes are balanced. \inlinecite{fatemi2024talk} notes that Erd{\"o}s-R{\'e}nyi (ER) graphs \cite{erdos59a} resulted in cycle check tasks where 82\% of the samples contained cycles. To address such issues of class imbalance, we enforce constraints on the number of samples which have any one answer. Table \ref{tab:task_data_constraints} summarizes our constraints for each task, with random accuracy indicating the constraint we place on answer distributions. Similar constraints are placed on both training and testing data, to ensure that class imbalance is addressed at each stage. 

\begin{table}[htb]
    \centering
    {\small
    \begin{tabular}{l l  r }
    Task & Ans. Type & Random Acc. \\
    \hline
    \rowcolor{Ivory2}Node Count (NC)& Numeric & 10\% \\
    \rowcolor{Ivory2}Node Degree (ND) & Numeric & 15\% \\
    \rowcolor{Ivory2}Depth-First Search (DFS) & Boolean & 50\% \\
    \rowcolor{Ivory2}Breadth-First Search (BFS) & Boolean & 50\% \\
    Edge Existence (EE) & Boolean & 50\% \\
    Edge Count (EC) & Numeric & 5\% \\
    Shortest Path (SP) & List & 15\%* \\
    Cycle Check (CC) & Boolean & 50\% 
    \end{tabular}}
    \caption{Data for each task is constrained such that randomly guessing the most common answer would achieve limited accuracy. *For SP, 15\% of answers have the same shortest path \textit{length}, and actual path lists are rarely repeated.}
    \label{tab:task_data_constraints}
\end{table}

\begin{table}[htb]
    \centering
    {\small
    \begin{tabular}{l c l l}

    Data Split & Seen Tasks? & Nodes & Edges\\
    \hline
    Train & - & [20, 100] & [0, 500]\\
    Validation & - & [20, 100]  & [0, 500]\\
    \rowcolor{Ivory2}In Dist. Test & \cmark & [20, 100] & [0, 500]\\
    \rowcolor{Ivory2}OOD Graph Size & \cmark & [140, 160] & [0, 500] \\
    \rowcolor{Ivory2}OOD Lengths & \cmark & [20, 100] & [0, 7800] \\
    \rowcolor{Ivory2}OOD Trees & \cmark & [20, 100] & [20, 100] \\
    \rowcolor{Ivory2}OOD Cycles & \cmark & [20, 120] & [20, 120] \\
    OOD Tasks & \xmark & [20, 100] & [0, 500] \\
    ProntoQA & \xmark & [14, 20]* & [12, 18]*\\
    LC-QuAD 2.0 & \xmark &  [5, 188]* & [4, 498]*\\
    \end{tabular}}
    \caption{Overview of test datasets. ``Seen Tasks?'' indicates whether the dataset's tasks are included in training. *Node and edge counts for PrOntoQA and LC-QuAD 2.0 are based on converting the question text and KG triples into graphs.}
    \label{tab:test_dataset_breakdown}
\end{table}

\subsection{Training Data}


We generate training and validation data for graphs containing 20 to 100 nodes, and bound the number of edges to 500. The constraint on the edges is motivated by memory limitations during training, ensuring that training could be performed on a single 80GB GPU. This also gives us an opportunity to test the generalization ability of our learned models to graphs with many more edges than were seen during training in a case where we controlled length.

We generate \textbf{3,000} training samples for each training tasks, for a total of \textbf{12,000} samples for multi-task training. 500 validation samples are also generated for each task. As in GraphQA, we select 7 models and algorithms to generate graphs; Erd{\"o}s-R{\'e}nyi, Barabási–Albert \cite{bagraph}, Stochastic Block Model \cite{SBMgraph}, Scale-Free graphs \cite{SFNgraph}, Star graphs, Path graphs, and Directed Acyclic graphs.

We train our LoRA and P-Tuning models for up to 20 epochs, and our Graph Token and Graph Token + Text models using a Relational Transformer \cite{diao2023relational} as the graph encoding model for up to 40 epochs. Details can be found in Appendix 3.

\subsection{Testing} 

We generate several variations of testing data for graph tasks to better assess the ability of various approaches to learning to solve graph tasks and generalizing to out of distribution (OOD) conditions and tasks. Unless otherwise noted, test graphs are generated using the same 7 graph structures as for training. Key statistics about the tasks and graph sizes for our test data can be seen in Table \ref{tab:test_dataset_breakdown}. 
Additional details surrounding data generation and statistics for our test data can be found in Appendix 4.

\subsubsection{In Distribution} 500 samples are generated for each task using the same procedure as for training data generation. 
\subsubsection{OOD Graph sizes} 500 samples are generated for each task for graphs containing 140 to 160 nodes.

\subsubsection{OOD Answer, Trace, and Path Lengths} 
500 test samples are generated for the BFS and DFS tasks, with no limit placed on the number of edges. Additionally, test data is generated such that sufficient test samples are produced for test cases with varying lengths of the answer (based on ground-truth answer token length, including the CoT), BFS/DFS trace lengths (i.e., the number of unique nodes visited during the search), and the shortest path length connecting nodes (for nodes which are reachable). These cases result in test data where the solution to the graph tasks are outside of the typical distribution seen for graphs during training -- for example, in our training data for DFS and BFS tasks, the search trace tended to be very short for cases where the nodes were not reachable, but that is not true of the test set.
\subsubsection{OOD Graph Structures (Trees and Cycles)} We experiment with two types of graph structures which were not generated by our default set of graph generation algorithms. One set is constructed by generating random power law trees\footnote{\url{https://tinyurl.com/4wpkpkwh}} containing 20 to 100 nodes (500 test samples per task). A second set is constructed for only the BFS and DFS tasks; \textit{cycle graphs} are generated similar to \inlinecite{abbe-et-al-neurips2024}. Cycle graphs consist of either 1 large cycle, for tests where nodes in question are reachable, or two smaller cycles, for tests where the nodes are not reachable. Graphs are constructed such that for a graph with $n$ nodes, the correct answer can be reached by following a BFS/DFS trace for $n/2$ steps. Cycle graphs are constructed for graphs containing 20 to 120 nodes.
\subsubsection{OOD Tasks} 500 test samples are generated for each of 4 graph tasks which were not used for training -- EE, EC, SP, and CC -- using the same procedure as our in-distribution test generation. 
Additionally, for further fine tuning we produce 30 additional training and validation samples for each task. Here, we evaluate generalization ability of each model on out of distribution tasks. Then, we  investigate the ability to learn new tasks with a relatively small set of training data. We compare both fine-tuned models, as well as training models from scratch using limited training data.

\subsubsection{Applicability to Question Answering}

We use PrOntoQA \cite{PrOntoQA}, which has a set of logical assertions (see Figure \ref{prontoqa:gen}), followed by an assertion.  The solution can be found by generating a graph between the atoms of the logical assertions and seeing if the atoms mentioned in the question are reachable.  We evaluate whether training on graph reasoning can help with answering questions on PrOntoQA.

We evaluate knowledge graph question answering (KGQA) using LC-QuAD 2.0 \cite{dubey2019lc}. Unlike the graph tasks seen during training, KGs are represented using triplets (\texttt{s, p, o}), while our training tasks graphs have no labeled edges. Additionally, KGQA questions are not answerable solely by graph reachability, as semantics of nodes and relations matter. We evaluate 475 boolean questions, because that is closest to the answer format at training, and curate a subset of the KG for each question from WikiData \cite{wikidata}.

See Appendix 5 for more details.

\section{Results}

\subsection{Preliminary Experiments}

In preliminary experiments, we observed the following:
\begin{itemize}
    \item Instruction tuning with reasoning steps is necessary to learn graph tasks for Graph Tokens and P-Tuning.
    \item A fixed token count for Graph Tokens regardless of graph size improved performance over a dynamic token count. 
    \item Some tasks not learned independently (such as Node Degree) can be learned in a multi-task setting.
    \item For Graph Tokens, including node labels in the prompt helps; initializing the GNN with embeddings of node labels as node features did not help.  
\end{itemize}

Details of these experiments are in Appendix 2 and 6.


\subsection{RQ1: Can LLMs Learn Graph Tasks?}

LLMs are able to learn to solve fundamental graph tasks quite well, as can be seen in Table \ref{tab:mainres_indist_and_oodist}. In this table and subsequent results, Random indicates the accuracy of predicting the most common answer class in the test data and 3-Shot Prompt indicates results using the base LLM with a 3-shot prompting strategy. We observe that all tasks are learned very effectively by all models for in-distribution graph sizes, with the exception of Graph Tokens on the Node Degree task. 

\begin{table}[h]
\centering
{\small
\begin{tabular}{l c c c c c}
Method & NC & ND & DFS & BFS \\
\hline
Random & .100 & .150 & .500 & .500 \\
\rowcolor{Ivory2}\multicolumn{5}{c}{In Distribution (smaller) Graphs} \\
\rowcolor{Ivory2}3-Shot Prompt & .492 & .290 & .382 & .202 \\
\rowcolor{Ivory2}LoRA & \textbf{1.00} & \textbf{1.00} & \textbf{.982} & \textbf{.996} \\
\rowcolor{Ivory2}P-Tuning & .998 & .996 & .978 & .988 \\
\rowcolor{Ivory2}Graph Tokens & .990 & .762 & .958 & .972 \\
\rowcolor{Ivory2}Graph Tokens + Text &  \textbf{1.00} & .992 & .978 & .988 \\
\multicolumn{5}{c}{Out of Distribution (larger) Graphs} \\
3-Shot Prompt & .366 & .212 & .340 & .330 \\
\textbf{LoRA} & \textbf{.758} & \textbf{1.00} & \textbf{.992} & \textbf{.994} \\

P-Tuning & .662 & .948 & .970 & .958 \\
Graph Tokens & .004 & .158 & .254 & .270 \\
Graph Tokens + Text & .478 & .874 & .852 & .762 \\

\end{tabular}}
\caption{Results for test cases compared by graph size.}
\label{tab:mainres_indist_and_oodist}
\end{table}

\subsection{RQ2a: Generalization of Graph Size and Structure}

Broadly, we observe that using LoRA together with multi-task training and instruction tuning, LLMs can learn to solve graph tasks and generalize across graph sizes and structures. Other approaches show greater performance degradation when applied different structures, especially cycle graphs. We also find that using Graph Tokens alone is particularly brittle when applied out of the training distributions, or out of training tasks.

\subsubsection{OOD Graph Size}
Results for OOD graph sizes can be seen in the lower half of Table \ref{tab:mainres_indist_and_oodist}. Graph Tokens' performance significantly drops in this setting. We expect that a strong contributing factor to this is the use of a fixed node count for Graph Tokens during training, where nodes past the first 100 would never have edges during training. Graph Tokens + Text remains somewhat more robust, but we observe that its performance is worse than P-Tuning.
All models show some decrease in performance on the node count task as all numeric answers for NC would be completely unseen during training over the 20-100 node graphs.  




\subsubsection{OOD Lengths}
For our OOD Lengths test set, we find that out of distribution answer lengths for BFS and DFS 
did not significantly affect outcomes of the model, with the exception of tests which investigated longer trace lengths. When applied to test cases with longer trace lengths, LoRA's accuracy remained above $.990$ for all tests, but the DFS performance of P-Tuning decreased from $.978$ to $.830$, Graph Tokens decreased from $.958$ to $.870$, and Graph Tokens + Text decreased from $.978$ to $.778$ -- similar decreases were seen for BFS. 
A likely explanation for this decrease is that the search trace required to determine when nodes were \textit{not} reachable tended to be quite short in our training data -- this test introduces many cases where long traces are needed to determine that nodes are not reachable. 
Full result tables for this set of experiments can be found in Appendix 7.

\subsubsection{OOD Graph Structures}

Table \ref{tab:mainres_tree} shows model performance for our two OOD graph structures, trees and cycle graphs. For the tree graphs, Graph Tokens once again shows some larger decrease in performance compared to the other methods. On the cycle graphs, LoRA continues to show very strong performance while our other models show a larger degree of degradation.

\begin{table}[h]
{\small
\centering
\begin{tabular}{l | c c c c | c c}
 & \multicolumn{4}{|c|}{ tree } & \multicolumn{2}{|c}{ cycle }\\
Method & NC & ND & DFS & BFS & DFS & BFS\\
\hline
\rowcolor{Ivory2}Random & .100 & .150 & .500 & .500 & .500 & .500\\
LoRA & \textbf{1.00} & \textbf{1.00} & \textbf{1.00} & \textbf{1.00} & \textbf{.998} & \textbf{.998}\\
P-Tuning & .998 & .982 & .966 & .996 & .927 & .882\\
Graph Tokens & .998 & .414 & .898 & .810 & .700 & .786 \\
Graph Tok+Text & \textbf{1.00} & .986 & .958 & .976 & .867 & .791\\
\end{tabular}}
\caption{Results for OOD graph structures}
\label{tab:mainres_tree}
\end{table}



\begin{figure}[htb]
    \centering
    \includegraphics[width=\linewidth]{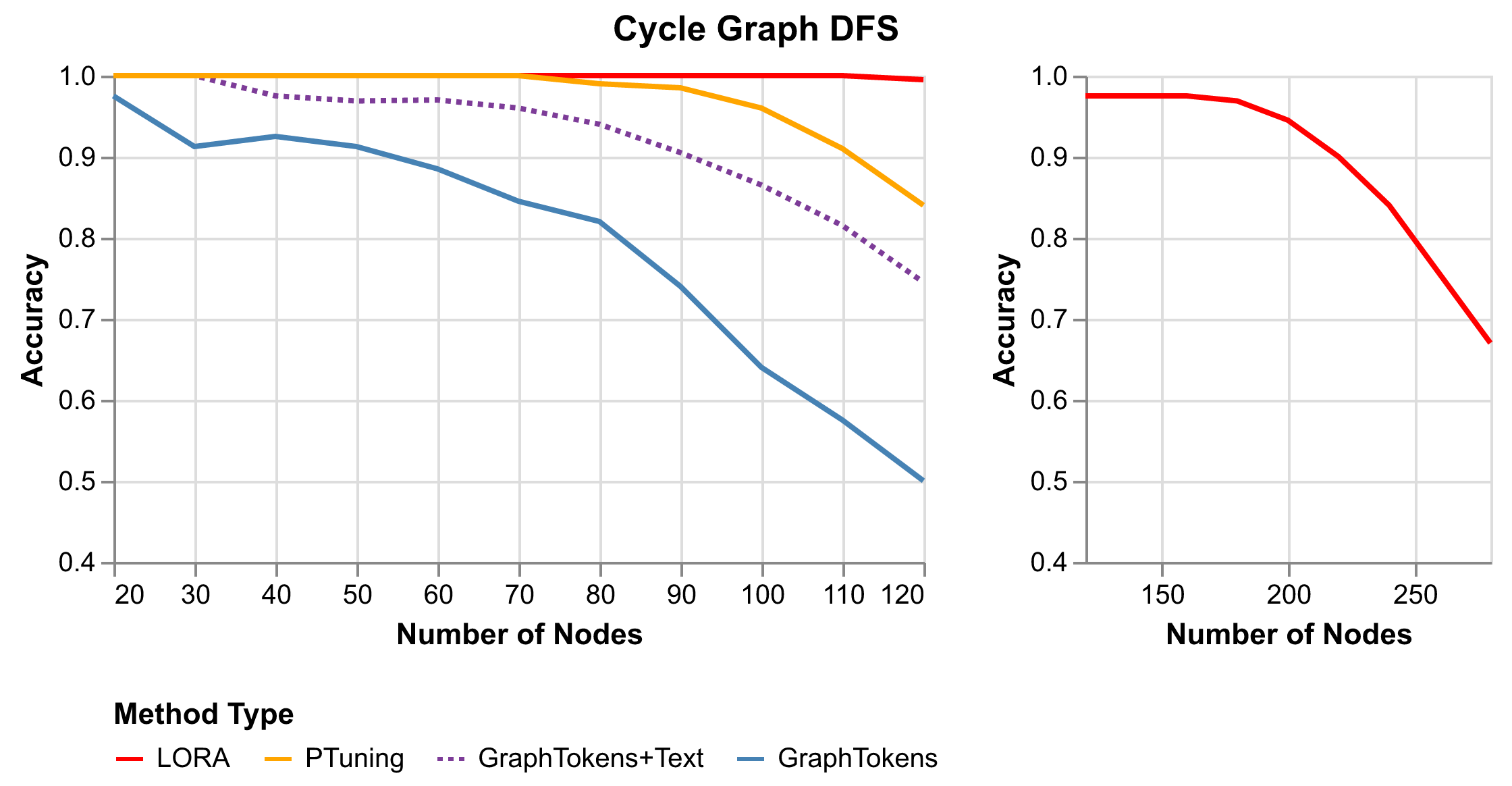}
    \caption{DFS accuracy v.s. graph size on cycle graphs.}
    \label{fig:test_cycv2}
\end{figure}

Figure \ref{fig:test_cycv2} highlights how models struggle to generalize to cycle graphs for the DFS task as the graph size and path length increase. 
Despite all other models having lower performance, LoRA continues to show strong performance over the cycle graphs. 


On the right side of Figure \ref{fig:test_cycv2}, we evaluate LoRA on much larger graphs, up to 280 nodes / 140-step DFS traces. Even for a 200 node graph -- twice the size seen during training -- LoRA shows strong performance. Eventually its accuracy drops closer to random, but our results show significant generalization beyond the training set both in terms of graph size and trace length, where 75\% of the training samples had traces with only 19 steps or fewer.

\subsection{RQ2b: Generalization to New Tasks}

Broadly, all models struggle to directly solve new tasks out-of-the-box, but the underlying reasoning steps they output often appear to be on track (due to formatting issues, it is difficult to assess the accuracy of these programmatically). Models often can still out-perform random and fewshot prompting baselines.

Table \ref{tab:ood_task_res_notrain} shows the results of our OOD tasks. We observe that our models all struggle to generalize to the cycle check task -- a key reason for this is that they tend to correctly follow the steps to perform a BFS to find the cycle, but they fail to stop after reaching the origin node which would close the cycle. As a result, they often continue to produce reasoning steps and never reach an answer, leading to performance below chance.

\begin{table}[!h]
\centering
{\small
\begin{tabular}{l c c c c}
Model & EE & SP & EC & CC \\
\hline
\rowcolor{Ivory2}Random & .500 & .150 & .050 & .500 \\
\rowcolor{Ivory2}3-Shot Prompt & .678 & .288 & .030 & \textbf{.526} \\
LoRA & .870 & .192 & \textbf{.122} & .348 \\
P-Tuning & .722 & .060 & .090 & .384 \\
Graph Tokens & .688 & .098 & .098 & .354 \\
Graph Tokens + Text & \textbf{.896} & \textbf{.398} & .090 & .388 \\
\end{tabular}}
\caption{Out of distribution task results. Besides 3-Shot Prompt, models are applied in a 0-shot manner after pretraining on our 4 training tasks.}
\label{tab:ood_task_res_notrain}
\end{table}

Table \ref{tab:ood_task_res_trained} compares the performance of models after performing multi-task training using 30 examples of each task. With only 30 training samples, the models can quickly learn new tasks, and the fine tuned models show better learning than training models from scratch on the same number of examples. EE is an exception, where training from scratch seemed to be comparable to fine tuning, but this task is also the easiest. For the Graph Tokens, we found that 30 samples were not sufficient to train the encoder model from scratch, and fine tuning also was less effective than for LoRA.

\begin{table}[!h]
\centering
{\small
\begin{tabular}{l c c c c}
Model & EE & SP & EC & CC \\
\hline
Random & .500 & .150 & .050 & .500 \\
\rowcolor{Ivory2}S LoRA & \textbf{.968} & .544 & .172 & .780 \\
\rowcolor{Ivory2}S P-Tuning & .756 & .232 & .086 & .484 \\
\rowcolor{Ivory2}S Graph Tokens & .246 & .102 & .002 & .492 \\
\rowcolor{Ivory2}S Graph Tokens + Text & .722 & .202 & .040 & .288 \\
FT LoRA & .966 & \textbf{.720} & \textbf{.198} & \textbf{.994} \\
FT P-Tuning & .742 & .222 & .054 & .436 \\
FT Graph Tokens & .904 & .250 & .070 & .694 \\
FT Graph Token + Text & .878 & .364 & .108 & .784 \\
\end{tabular}}
\caption{Out of distribution task results after training the model on 30 task samples. FT indicates fine-tuning the models after training on our 4 training tasks, S indicates base model trained from scratch only using the new task samples.}
\label{tab:ood_task_res_trained}
\end{table}

\subsection{RQ2c: Question Answering Tasks}

While Graph Tokens shows some  generalization to new graph tasks, we found that their performance collapsed when applied to question answering tasks with unfamiliar inputs. Additionally, Graph Tokens requires a properly formatted graph to serve as input for the encoding, which may not be easily producible for many question answering tasks. As a result, our analysis on these tasks are limited to the LoRA and P-Tuning models.

\subsubsection{ProntoQA}

\begin{figure}[h]
\centering
\begin{mdframed}[backgroundcolor=Ivory1]
\scriptsize
Here are a set of statements: 

Every petun is not green. Petuns are gretons. Every greton is not smart. Gretons are not luminous.

Create a graph from these facts. Format it so it looks like this:

Below is a graph of nodes, where x $\to$ y, z means the graph has edges (x, y) and (x, z). 

\xmltag{graph}

petun $\to$ not green, greton 

greton $\to$ not smart, not luminous

\xmltag{graph}

Here are some new statements (PrOntoQA problem):

Every tumpus is not angry. Tumpuses are rompuses. Every numpus is not bright. Rompuses are not luminous... Stella is a yumpus.  Create a graph with these facts, in the same format.
\end{mdframed}
\caption{Prompt for graphs from PrOntoQA problems}
\label{prontoqa:gen}
\end{figure}

We re-structured the PrOntoQA task into two tasks - 1. \texttt{id\_graphs} where we manually parsed logical statements into graphs in the format used for training , and 2. \texttt{generated} where we prompted the base model to produce a graph that we then used with the trained models for reasoning (see Figure~\ref{prontoqa:gen} for an example prompt used for graph generation). Crucially, the graph structure in \texttt{generated} is \textit{different} from those used in training.  Table \ref{tab:prontoqa_res} shows results on PrOntoQA, compared with the case when no training was involved.  In a now familiar pattern, both LoRA and P-Tuning demonstrated generalization with LoRA showing much greater generalization than P-Tuning.  

\begin{table}[h]
\centering
{\small
\begin{tabular}{l l c}
Model & Prompt Type & Accuracy \\
\hline
Base & original & .472 \\
Base & id\_graphs & .470 \\
Base & generated & .546 \\
\rowcolor{Ivory2}LoRA & id\_graphs & \textbf{.998} \\
\rowcolor{Ivory2}LoRA & generated & .774 \\
P-Tuning & id\_graphs & .608 \\
P-Tuning & generated & .580 \\
\end{tabular}}
\caption{Results for the PrOntoQA task.}
\label{tab:prontoqa_res}
\end{table}

\subsubsection{LC-QuAD 2.0}



Table \ref{tab:lcquad_res} shows results on LC-QuAD. In the prompt type, 
\texttt{triples} indicates that the KG subgraph was fed in as triplets, i.e. \texttt{(subj, pred, obj)}, into the question prompt. \texttt{id\_triples} indicates that we converted the triplets to be somewhat similar to our training data, with nodes being given IDs and expressing \texttt{pred, obj} pairs as a single node label. We provide accuracy in terms of all boolean questions, simply as ``Accuracy'', as well as accuracy over 156 questions which did not ask questions about numeric values as ``N.N. Accuracy'' (non-numeric accuracy).

\begin{table}[h]
\centering
{\small
\begin{tabular}{l l r r}
Model & Prompt Type & Accuracy & N.N. Accuracy \\
\hline
Random & - & .522 & .616 \\
\rowcolor{Ivory2}Base & triples & .707 & .519\\
\rowcolor{Ivory2}Base & id\_triples & .507 & .263 \\
LoRA & triples & .625 & .462\\
LoRA & id\_triples & .621 & .641 \\
S LoRA & id\_triples & .703 & .474 \\
FT LoRA & id\_triples & \textbf{.747} & \textbf{.660} \\
\rowcolor{Ivory2}P-Tuning & triples & .291 & .244\\
\rowcolor{Ivory2}P-Tuning & id\_triples & .404 & .462 \\
\rowcolor{Ivory2}S P-Tuning & id\_triples & .602 & .423 \\
\rowcolor{Ivory2}FT P-Tuning & id\_triples & .528 & .449 \\
\end{tabular}}
\caption{Results for the LC-QuAD task. FT indicates fine-tuning the models after training on our 4 training tasks, S indicates base model trained from scratch only using the new task samples.}
\label{tab:lcquad_res}
\end{table}

We observe several key elements which appear to make our trained models struggle to generalize to KGQA. First, in addition to the KG size being OOD compared to our training graphs, introducing relation information as triplets is quite different from training where graphs had no edge labels. Second, many boolean questions for KGQA cannot be answered by just following paths or counting nodes, as was done in our training tasks. For example, questions may require checking multiple paths like ``Did Pope Paul VI work in both Rome and Munich?'', or comparing values like ``Is the luminosity of the Alpha Andromedae less than 240?''. 
Third, because our training tasks use integers to indicate node IDs, questions which include numeric values appear to confuse the LLM. 
To examine the impact of this last case, our ``N.N. Accuracy'' shows that when we disregard such questions, LoRA has better performance compared to the base model, showing some generalization.

We also  explore whether fine tuning on 30 examples can enable better performance. Fine tuning does help to improve our models' ability to understand the KG structure, despite the fact that LC-QuAD is challenging due to the need to go well beyond the graph tasks it was trained on. Additionally, fine tuning our pretrained LoRA shows better performance than training a model from scratch.

\subsection{RQ3: Tradeoffs and Limitations}

Graph Tokens with no added text can learn many tasks, but they are less effective than methods which rely on graph text and show poorer OOD generalization. Graph Tokens have a major advantage in input token count, especially for graphs with many edges. OOD tasks should be a future focus, as reduced input token count can improve the efficiency of LLMs.

P-Tuning demonstrates that LLMs are capable of using prompt changes to solve graph tasks. In many cases, we saw P-Tuning match or beat the Graph Tokens + Text approach, despite P-Tuning using a static set of tokens.

P-Tuning and Graph Tokens are both more portable across different LLMs. LoRA shows the strongest performance, but it is specialized to the LLM in our experiments; a Graph Token encoder can be trained with one LLM and applied to another, and P-Tuning can be adapted by training a projection layer to change the embedding size of the learned tokens. We briefly explore this idea in Appendix 8.

\section{Conclusion}


LLMs can be trained to reason on graphs, and show good generalization without the need for special graph encoding models.  These results suggest that reasoning over graph tasks which are more open ended (for example, more complex KGQA tasks) and training a model to learn \textit{when} to solve a problem as a graph task, and perhaps treating the actual graph task as a tool, rather than trying to always use the LLM directly to compute over the graph, may be valuable to explore.  


\bibliography{aaai2026}

\clearpage

\section{Appendix 1: Task Details and Example Text}

\subsection{Graph Representation}

Figure \ref{apfig:graph_rep_example} shows the full textual representation of a graph, which is prepended to the input question for all tasks and models, except for the Graph Tokens approach where we omit the portion from ``The edges in G are:'' onwards. 

\begin{figure}[!h]
\centering
\begin{tcolorbox}
\scriptsize
In a directed graph, the mapping of node IDs to their labels is given by a dictionary. Edges are represented as i -$>$ j,k means that there is an edge from node i to node j, and another edge from node i to k. G describes a graph among nodes with the following mapping from IDs to labels:

\{0: Phasmatida,
 
1: mandruka,

2: eulogy,

3: benzoiodohydrin,

4: Krishnaitic\} 

The edges in G are: 

0 -$>$ 1, 2

1 -$>$ 3

2 -$>$ 1, 4

4 -$>$ 0

\end{tcolorbox}
\caption{Example representation of a graph and question.}
\label{apfig:graph_rep_example}
\end{figure}

All subsequent example texts shown for the tasks in this section will use the above graph as input.

\subsection{Task Examples}

The following figures will show example questions as well as answer CoTs for each task. During inference, only text that comes before the \texttt{$<$think$>$} tag are input to the model. Each question has 5 alternate phrasings to try to reduce overfitting, including re-ordering the nodes in reachability questions (e.g. ``can NodeA reach NodeB'' versus ``is NodeB reachable from NodeA'').

\begin{figure}[!h]
\centering
\begin{tcolorbox}
\scriptsize
Q: How many nodes does the graph contain?

A: 

$<$think$>$To count the nodes in the graph, gather the nodes into a list:

The list of nodes is [0,1,2,3,4]

The number of nodes is the length of the list 5$<$/think$>$$<$answer$>$5$<$/answer$>$
\end{tcolorbox}
\caption{Node Count task.}
\end{figure}

\begin{figure}[!h]
\centering
\begin{tcolorbox}
\scriptsize
Q: How many other nodes can be reached by "eulogy" in a single step?

A: 

$<$think$>$From the label mapping, we know that eulogy maps to 2. Counting neighbors of 2:

(2, 1) count:1

(2, 4) count:2

$<$/think$>$$<$answer$>$2$<$/answer$>$
\end{tcolorbox}
\caption{Node Degree task.}
\end{figure}

\begin{figure}[!h]
\centering
\begin{tcolorbox}
\scriptsize
Q: Can a depth first search be used to reach "benzoiodohydrin" from "eulogy"?

A:

$<$think$>$Starting depth first traversal from eulogy to see if benzoiodohydrin is reachable. From the label mapping, eulogy maps to 2, and benzoiodohydrin maps to 3.

Starting depth first traversal from 2. Backtrack once there are no more new neighbors to visit or 3 is reached.

2 -$>$ 1 -$>$ 3

Reached 3

$<$/think$>$$<$answer$>$Yes$<$/answer$>$
\end{tcolorbox}
\caption{Depth-First Search task.}
\end{figure}

\begin{figure}[!h]
\centering
\begin{tcolorbox}
\scriptsize
Q: Can a breadth first search be used to reach "benzoiodohydrin" from "eulogy"?

A:

$<$think$>$Starting breadth first traversal from eulogy to see if benzoiodohydrin is reachable. From the label mapping, eulogy maps to 2, and benzoiodohydrin maps to 3. Visit all new neighbors of nodes in the current depth's queue, skipping visited nodes.

Queue: [2]

2 -$>$ 1, 4

Queue: [1, 4]

1 -$>$ 3

Reached 3

$<$/think$>$$<$answer$>$Yes$<$/answer$>$
\end{tcolorbox}
\caption{Breadth-First Search task.}
\end{figure}

\begin{figure}[!h]
\centering
\begin{tcolorbox}
\scriptsize
Q: Is there an edge from "eulogy" to "benzoiodohydrin"?

A: 

$<$think$>$To check if the edge exists, eulogy is mapped to 2, and benzoiodohydrin is mapped to 3.  Check if 2-$>$3:

2 has edges to [4,1]

$<$/think$>$$<$answer$>$No$<$/answer$>$
\end{tcolorbox}
\caption{Edge Existence task.}
\end{figure}

\begin{figure}[!h]
\centering
\begin{tcolorbox}
\scriptsize
Q: How many total edges are there in the entire graph? Write out a list of all edges in the graph, then answer the size of the list.

$<$think$>$To count edges, count the edges from each node in the graph:

The list of edges is: [0-$>$1,0-$>$2,1-$>$3,2-$>$4,2-$>$1,4-$>$0]
The answer is the length of the list 6

$<$/think$>$$<$answer$>$6$<$/answer$>$
\end{tcolorbox}
\caption{Edge Count task.}
\end{figure}

\begin{figure}[!h]
\centering
\begin{tcolorbox}
\scriptsize
Q: What is the shortest path from "eulogy" to "benzoiodohydrin"? Use BFS to find the shortest path, then output the path as a list. If there is no path output an empty list.

A: 

$<$think$>$Starting breadth first traversal from eulogy to benzoiodohydrin to find the shortest path. From the label mapping, eulogy maps to 2, and benzoiodohydrin maps to 3. Visit all new neighbors of nodes in the current depth's queue, skipping visited nodes.

Queue: [2]

2 -$>$ 1, 4

Queue: [1, 4]

1 -$>$ 3

Reached 3

Work backwards from reached 3 to nodes that have an backwards edge to 2 The shortest path is therefore:

$<$/think$>$$<$answer$>$[2, 1, 3]$<$/answer$>$
\end{tcolorbox}
\caption{Shortest Path task.}
\end{figure}

\begin{figure}[!h]
\centering
\begin{tcolorbox}
\scriptsize
Q: Does the graph contain any cycles starting from "Krishnaitic"? Check if there is a path from "Krishnaitic" to itself.

A: 

$<$think$>$Starting depth first traversal from Krishnaitic to see if Krishnaitic is reachable. From the label mapping, Krishnaitic maps to 4, and Krishnaitic maps to 4.

Starting depth first traversal from 4. Backtrack once there are no more new neighbors to visit or 4 is reached.

4 -$>$ 0 -$>$ 1 -$>$ 3

0 -$>$ 2 -$>$ 1

2 -$>$ 4

Reached 4

$<$/think$>$$<$answer$>$Yes$<$/answer$>$
\end{tcolorbox}
\caption{Cycle Check task.}
\end{figure}

\clearpage
\section{Appendix 2: Comparison of Model Performance using CoT}

As part of our preliminary exploration, we conducted experiments to validate the performance of various models when we trained the model to output a chain-of-thought reasoning process to answer questions versus when we trained the models to directly output an answer. For this set of experiments, we used a set of graphs containing only 5 to 20 nodes, similar to many other prior works. 

\begin{table}[!htb]
    \centering
    \begin{tabular}{l l l l l l}
    Model & CoT? & NC & ND & DFS & BFS \\
    \hline
    LoRA & \xmark & 1.0 & .99 & .96 & .97 \\
    LoRA & \cmark & 1.0 & 1.0 & .99 & 1.0\\

    \rowcolor{Ivory2}P-Tuning & \xmark & 1.0 & .31 & .75 & .81 \\
    \rowcolor{Ivory2}P-Tuning & \cmark & 1.0 & .99 & .98 & 1.0 \\

    Graph Tokens & \xmark & .96 & .36 & .85 & .87 \\
    Graph Tokens & \cmark & .96 & .91 & .99 & .99 \\
    \end{tabular}
    \caption{Model performance when trained with or without CoT.}
    \label{apptab:cot_nocot}
\end{table}

As can be seen in Table \ref{apptab:cot_nocot}, across all methods using CoT leads to better test performance. In particular, P-Tuning and Graph Tokens see significant performance decreases for the Node Degree task when no CoT is used, despite the fact that the task should be very easy. 

LoRA's performance looks quite good even without CoT. To investigate this point further, we also conducted additional experiments with LoRA on our main experiments' graph sizes (graphs containing 20 to 100 nodes and 500 or fewer edges). Surprisingly, we find that LoRA without CoT still appears to show very good performance, shown in Table \ref{apptab:lora_nocotbig}.

\begin{table}[!htb]
    \centering
    \begin{tabular}{l l l l l l}
    Model & CoT? & NC & ND & DFS & BFS \\
    \hline
    LoRA & \xmark & .99 & .99 & .96 & .95 \\
    LoRA & \cmark & 1.0 & 1.0 & .98 & .99\\
    \end{tabular}
    \caption{LoRA performance with or without CoT, trained and tested on larger graphs}
    \label{apptab:lora_nocotbig}
\end{table}

However, without CoT, LoRA is \textbf{not} able to generalize so well. The lack of generalization is especially apparent when testing on the cycle graphs, shown in Table \ref{apptab:lora_nocot_cyc}. Here, the model clearly is not able to learn how to generalize to novel graph structures or longer path lengths. On the other hand, our LoRA model trained with CoT continues to show incredibly strong generalization to this graph structure.

\begin{table}[!htb]
    \centering
    \begin{tabular}{l l l l l l}
    Model & CoT? & DFS & BFS \\
    \hline
    LoRA & \xmark & .48 & .48 \\
    LoRA & \cmark & .99 & .99 \\
    \end{tabular}
    \caption{LoRA performance with or without CoT, tested on cycle graphs.}
    \label{apptab:lora_nocot_cyc}
\end{table}

\section{Appendix 3: Model Parameters and Architecture}

\subsection{LoRA}

We make use of the PEFT library's implementation of LoRA\footnote{https://huggingface.co/docs/peft/en/package\_reference/lora} for our experiments, mostly using default parameters. We select the target modules \texttt{["q\_proj", "k\_proj", "v\_proj", "o\_proj"]}, rank \texttt{r=16}, alpha \texttt{lora\_alpha=32}, and dropout \texttt{lora\_dropout=0.1}. 

LoRA is trained for up to 20 epochs, with early stopping based on loss on the validation set with a patience of 5 epochs. In practice, LoRA often achieved its best validation loss around epoch 10-15.

\subsection{P-Tuning}

Similarly, we make use of PEFT's implementation of P-Tuning\footnote{https://huggingface.co/docs/peft/en/package\_reference/p\_tuning}, again mostly using default parameters provided by the PEFT library. The only specialized parameter we set is the number of virtual tokens, \texttt{num\_virtual\_tokens=400}. This number reflects the number of tokens used by our Graph Tokens and Graph Tokens + Text models, and is set to make the results of P-Tuning more directly comparable to these competing approaches. 

As with LoRA, P-Tuning is trained for up to 20 epochs, with early stopping based on loss on the validation set with a patience of 5 epochs. P-Tuning also tended to reach its best validation loss around epoch 10-15.

\subsection{Graph Tokens}

In the Graph Tokens approach presented by Perozzi et al. (2024), the graph encoding model, node or edge aggregation, and tokenization varied for each of task. In our case, we want to explore the abilities of a Graph Token approach using a single model applied to different tasks.

After exploring several graph encoding methods proposed by the original paper, we opted to use the Relational Transformer (Diao and Loynd 2023) (RT) as our main graph encoding layer. While applied in a very different setting than our paper, RT showed very strong performance compared to other graph neural network architectures at learning to solve graph reasoning tasks, and additionally demonstrated adaptability to OOD graph sizes. RT also has the benefit of applying attention while maintaining $n^2$ computational complexity for a graph with $n$ nodes, unlike many other works which require $e^2$ over the number of edges $e$ in a graph.

We re-implemented RT into \texttt{torch} modules, and construct our graph encoding model using 6 RT layers (explored 1 to 12 layers), node embedding sizes of 384 (explored embedding sizes from 64 to 768), hidden sizes of 64 (explored sizes of 32 to 128), 12 attention heads (explored number of attention heads from 2 to 12), and a dropout of 0.3.

Another important element of the Graph Token architecture is our choice of node features to serve as input to the graph encoder. We experimented with three approaches for node features: producing sentence embeddings of node labels, Laplacian position embeddings, and Sinusoidal position embeddings. We found that simply using Sinusoidal position embeddings was the most effective in our experiments. 

Lastly, for our graph into to the graph encoder model, we designed our model such that we always input a fixed size graph, regardless of the task at hand, and similarly output a fixed number of graph tokens. In addition to using a fixed graph size enabling better parallelization during training, we found that the model struggled to effectively learn all of our tasks when we allowed the input and output graph sizes to vary. In our main experiments, we set a fixed graph size of 200 nodes, and output 400 total embeddings to serve as our graph tokens (200 node embeddings, and 200 edge embeddings, computed by aggregating the sum of the adjacency matrix output by our graph encoder model). Using a fixed graph size, in practice each problem only changes the adjacency matrix input to the graph encoder.

\begin{figure}
    \centering
    \includegraphics[width=\linewidth]{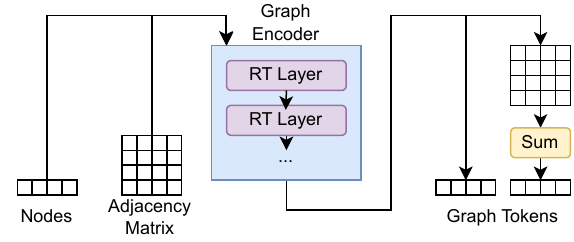}
    \caption{High level overview of the graph encoder.}
    \label{appfig:graphencoder}
\end{figure}

A high level overview of the graph encoder is shown in Figure \ref{appfig:graphencoder}. The graph encoder takes as input nodes and an adjacency matrix, and similarly outputs updated representations of nodes and an adjaency matrix -- inner RT layers take the same input and outputs. A projection layer is used to resize the nodes and adjacency matrix at both the input and output of the encoder.

For training, we used 40 epochs for Graph Tokens, again with early stopping based on validation loss. We typically found that 20 epochs were not sufficient for training the Graph Token model, and early stopping often triggered around 30-35 epochs.

\subsection{Graph Tokens + Text}

Our Graph Tokens + Text method used the exact same training procedure and model architecture as the Graph Tokens method, except that the input question text also included the edge information (i.e., ``The edges in G are: ...''). The choice of token count used for this model and P-Tuning make them affect the LLM in very similar ways, with the main difference being that P-Tuning inputs a static set of tokens for every question while Graph Tokens + Text outputs a different set of tokens for each question.

While the Graph Token + Text approach has an additional encoder, we find that the training time per epoch is nearly identical to that of our P-Tuning model. In practice, the computational cost for training and inference is dominated by the LLM. The Graph Token model without the additional text has the fastest training due to its reduction in input tokens.

\section{Appendix 4: Data Statistics}

We used 7 random graph generation algorithms and models for our training and test data generation, except for our OOD graph structures. When randomly generating our data, we also applied different weightings to how likely it was to select a given algorithm. The reason for this was that some graph structures are more interesting than others, and many graph structures were more likely to result in data which did not fit our desired distribution for certain tasks. Additionally, in the process of data generation, if a given answer already reached our answer distribution constraint, we would not include the generated data in our dataset -- this resulted in some algorithms producing more data than others.

Table \ref{apptab:alg_repr} shows the presence of each graph type within our training set, both in terms of absolute count and percentage. Each training task has 3,000 samples each, for a total of 12,000 graphs. We can see that some graph types are much more likely to occur in our data than others -- nearly 60\% of our training data uses scale-free network graphs. This is caused in part due to some graph structures being much more likely to have certain properties which would lead to more similar answers. For example, Path graphs would always have a node degree of 1, and therefore would no longer be added to the dataset after enough training samples with an answer of 1 have been generated.

\begin{table}[!htb]
    \centering
    \begin{tabular}{l r r}
    Graph Type & Count & Percentage \\
    \hline
    ER Graph & 1,811 & 30.2\% \\
    BA Graph & 2,060 & 34.3\% \\
    SBM Graph & 1,688 & 14.1\% \\
    SFN Graph & 7,132 & 59.4\% \\
    DAG & 1,398 & 23.3\% \\
    Star Graph & 2,428 & 20.2\%\\
    Path Graph & 1,107 & 18.4\% \\
    \end{tabular}
    \caption{Distribution of graph types in training data.}
    \label{apptab:alg_repr}
\end{table}

\begin{table}[!htb]
    \centering
    \footnotesize
    {\small
    \begin{tabular}{l r r r r r}

    Data Split & Q1 & Median & Mean & Q3 & Max \\
    \hline
    \rowcolor{Ivory2}Train & 30.0 & 43.0 & 50.9 & 72.0 & 100 \\
    
    In Dist. Test & 30.0 & 43.0 & 51.4 & 74.0 & 100 \\
    OOD Graph Size & 145.0 & 150.0 & 149.8 & 155.0 & 160 \\
    OOD Ans. Len. & 44.0 & 74.0 & 66.3 & 88.0 & 100 \\
    OOD Trace Len. & 42.0 & 67.0 & 64.3 & 87.0 & 100 \\
    OOD Path Len. & 42.0 & 65.0 & 63.3 & 85.0 & 100 \\
    OOD Trees & 38.0 & 61.0 & 60.9 & 84.0 & 99 \\
    OOD Cycles & 40.0 & 70.0 & 70.0 & 100.0 & 120 \\
    OOD Tasks & 30.0 & 43.0 & 51.2 & 75.0 & 100 \\
    \end{tabular}}
    \caption{Overview of node count statistics for training and test datasets. }
    \label{apptab:data_nodecounts}
\end{table}

\begin{table}[!htb]
    \centering
    \footnotesize
    {\small
    \begin{tabular}{l r r r r r}

    Data Split & Q1 & Median & Mean & Q3 & Max \\
    \hline
    \rowcolor{Ivory2}Train & 63.0 & 114.0 & 156.3 & 222.0 & 500 \\
    
    In Dist. Test & 64.0 & 118.0 & 158.1 & 223.0 & 500 \\
    OOD Graph Size & 152.0 & 221.0 & 208.7 & 243.0 & 500 \\
    OOD Ans. Len. & 57.0 & 87.5 & 202.9 & 142.2 & 5844 \\
    OOD Trace Len. & 84.0 & 234.5 & 744.7 & 983.2 & 7773 \\
    OOD Path Len. & 55.0 & 83.0 & 186.5 & 123.2 & 5844 \\
    OOD Trees & 37.0 & 60.0 & 59.9 & 83.0 & 98 \\
    OOD Cycles & 40.0 & 70.0 & 70.0 & 100.0 & 120 \\
    OOD Tasks & 57.0 & 95.0 & 136.0 & 167.0 & 500 \\
    \end{tabular}}
    \caption{Overview of edge count statistics for training and test datasets. }
    \label{apptab:data_edgecounts}
\end{table}

Here we show some more detailed statistics surrounding our generated datasets. Table \ref{apptab:data_nodecounts} shows the node counts in the data, and Table \ref{apptab:data_edgecounts} shows the edge counts. Both are presented in terms of the first quartile, median, mean, third quartile, and max node or edge counts seen in each dataset. 

Additionally, we present information surrounding trace lengths for the DFS task for each of our datasets in Table \ref{apptab:dfs_trace_lengths}. Trace lengths are computed for reachability questions regardless of whether the nodes are reachable, as the trace length indicates how many steps the CoT needs to search before determining the correct solution.

\begin{table}[!htb]
    \centering
    \footnotesize
    {\small
    \begin{tabular}{l r r r r r}
    Data Split & Q1 & Median & Mean & Q3 & Max \\
    \hline
    \rowcolor{Ivory2}Train & 4.0 & 10.0 & 14.4 & 20.0 & 91 \\
    
    In Dist. Test & 5.0 & 11.0 & 14.6 & 19.0 & 77 \\
    OOD Graph Size & 8.0 & 20.0 & 31.2 & 49.0 & 139 \\
    OOD Ans. Len. & 10.0 & 20.0 & 27.2 & 42.0 & 88 \\
    OOD Trace Len. & 9.0 & 19.0 & 21.1 & 30.0 & 90 \\
    OOD Path Len. & 12.0 & 22.0 & 25.8 & 36.0 & 92 \\
    OOD Trees & 1.0 & 8.0 & 15.4 & 23.0 & 90 \\
    OOD Cycles & 20.0 & 35.0 & 35.0 & 50.0 & 60 \\
    \end{tabular}}
    \caption{DFS task trance length statistics for each data split. Q1 and Q3 indicate the first quartile and third quartile, respectively.}
    \label{apptab:dfs_trace_lengths}
\end{table}

Several key observations may be made which can provide further insight into our experimental results surrounding DFS. In particular, the cycle graphs tend to have much larger traces on average due to the very deliberate manner in which they were constructed. This likely leads to exposing the models to more test cases with traces that were poorly represented during training. In contrast, even for our larger graph sizes or test data which aimed to produce longer answers, the average trace lengths tended to hover around 20.0.

\section{Appendix 5: PrOntoQA and LC-QuAD Data Creation Details and Example Text}

\subsection{PrOntoQA}

Figures \ref{appfig:prontoqa_og} shows a PrOntoQA question in its original form. A number of assertions -- using nonsensical words -- are made, and a final assertion is asked as part of the question. The goal here is to determine if the final assertion is supported by the other statements.

\begin{figure}[h]
\centering
\begin{mdframed}[backgroundcolor=Ivory1]
\scriptsize
Jompuses are not shy. Jompuses are yumpuses. Each yumpus is aggressive. Each yumpus is a dumpus. Dumpuses are not wooden. Dumpuses are wumpuses. Wumpuses are red. Every wumpus is an impus. Each impus is opaque. Impuses are tumpuses. Numpuses are sour. Tumpuses are not sour. Tumpuses are vumpuses. Vumpuses are earthy. Every vumpus is a zumpus. Zumpuses are small. Zumpuses are rompuses. Max is a yumpus. 

Question: Is the following statement true or false? Max is sour.
\end{mdframed}
\caption{Original question}
\label{appfig:prontoqa_og}
\end{figure}

A version of these questions which was automatically generated using our trained LLM is shown in Figure \ref{appfig:prontoqa_gen}. Here, the ``graph'' which is generated looks quite unlike our training graphs or a proper graph at all. However, we find that producing a graph-like structure like this leads to better performance with our trained models.

\begin{figure}[h]
\centering
\begin{mdframed}[backgroundcolor=Ivory1]
\scriptsize
Below is a graph of nodes, where x $\to$ y, z means the graph has edges (x, y) and (x, z). 

jompus $\to$ not shy,yumpus 
yumpus $\to$ aggressive,dumpus 
dumpus $\to$ not wooden,wumpus 
wumpus $\to$ red,impus 
impus $\to$ opaque,tumpus 
numpus $\to$ sour 
tumpus $\to$ not sour,vumpus 
vumpus $\to$ earthy,zumpus 
zumpus $\to$ small,rompus 
max $\to$ yumpus 

Answer the question: Is the following statement true or false? Max is sour..  Make sure to ensure that you do BFS on the graph using \xmltag{think}..\xmltag{/think} tags. Enclose answer in \xmltag{answer}..\xmltag{/answer}.  Answer with only A or B - given t\
he following options ['A) True', 'B) False'].

\end{mdframed}
\caption{Prompt using a graph generated from a pretrained model to answer questions}
\label{appfig:prontoqa_gen}
\end{figure}

Lastly, Figure \ref{appfig:prontoqa_id} shows a PrOntoQA question which was reformatted by us to resemble the graph formatting used during model training. Perhaps unsurprisingly, this approach leads to be best performance by models. 

\begin{figure}[h]
\centering
\begin{mdframed}[backgroundcolor=Ivory1]
\scriptsize
In a directed graph, the mapping of node IDs to their labels is given by a dictionary. Edges are represented as i $\to$ j,k means that there is an edge from node i to node j, and another edge from node i to k. G describes a graph among nodes with the following mapping from IDs to labels: \{0:jompus, 1:not\_shy, 2:yumpus, 3:aggressive, 4:dumpus, 5:not\_wooden, 6:wumpus, 7:red, 8:impus, 9:opaque, 10:tumpus, 11:numpus, 12:sour, 13:not\_sour, 14:vumpus, 15:earthy, 16:zumpus, 17:small, 18:rompus, 19:max\} 

The edges in G are: 
0 $\to$ 1,2 
2 $\to$ 3,4 
4 $\to$ 5,6 
6 $\to$ 7,8 
8 $\to$ 9,10 
11 $\to$ 12
10 $\to$ 13,14 
14 $\to$ 15,16 
16 $\to$ 17,18 
19 $\to$ 2

Do a BFS on the graph to help answer the question: Is the following statement true or false? Max is sour.  Pay attention to the word \"not\" in question when mapping it to a node in the graph.  Answer with only A or B - given the following options ['A) True', 'B) False'].  Answer True if there is a path, and False if there is no visited nodes remain.

\end{mdframed}
\caption{Prompt using an id graph generated by parsing the question to answer questions}
\label{appfig:prontoqa_id}
\end{figure}

\subsection{LC-QuAD 2.0}

As a KGQA dataset, LC-QuAD mainly aims to evaluate the ability of models to learn how to generate SPARQL queries to answer natural language questions. In our case, we are interested in repurposing it to analyze the underlying KG and produce answers. 

Given that LC-QuAD has a categorization of its data by the type of solution, we selected all \texttt{boolean} type questions from its test set to use in our tests. We run each question's ground truth SPARQL query against the public WikiData SPARQL endpoint\footnote{https://query.wikidata.org/} and retain all questions which produce a valid answer. Additionally, we set aside 30 questions to use as additional training data, for which we manually produce CoT that resembles the CoT seen by our models for training.

To make the data suitable for our evaluation, we curated a subgraph of WikiData surrounding each question's ground truth SPARQL query and use it as our input graph. Our evaluation task using LC-QuAD therefore is directly converted into a boolean task, where we provide the KG to the model in the form of a prompt, ask a natural language question, and expect a True or False answer. To collect a subgraph for each question, we first parse each SPARQL query to identify the key entities and properties needed to answer the question. We then query WikiData to collect a 3-hop neighborhood surrounding each such entity. Lastly, we sub-sample this subgraph further by sampling 5, 4, and 3 neighbors which are 1, 2, and 3 hops away, respectively. Labels for each entity and property in this subgraph are also collected from WikiData -- we remove any entities with missing labels. We randomly sample the subgraph such that the original ground truth SPARQL query can still properly be answered. This curated subgraph is used as our input graph for each question.

In our experiments, we format the KG in terms of triples and as an id\_graph. Triples feed in the KG triples simply as \texttt{(subject, predicate, object)}, very similar as standard KG serialization formats such as NTriples. id\_graph aims to convert the structure into a format which is more similar to what our models see during training, in order to investigate how well they might generalize. Examples of these two input formats can be seen in Figure \ref{appfig:triple_lcquad} and \ref{appfig:idgraph_lcquad}

\begin{figure}[!h]
\centering
\begin{mdframed}[backgroundcolor=Ivory1]
\scriptsize
Below is a graph of nodes represented as triplets, where a triple (s, p, o) indicates that node s is connected to node o by the relation p.

('LuAZ-967', 'wheelbase', '1800')

('amphibious vehicle', 'described by source', 'Armenian Soviet Encyclopedia, vol. 1')

('LuAZ-967', 'ride height', '280')

('LuAZ-967', 'length', '3682')

('LuAZ-967', 'height', '1580')

('LuAZ-967', 'subclass of', 'amphibious vehicle')

('amphibious vehicle', 'subclass of', 'watercraft')

('LuAZ-967', 'width', '1712')

('Soviet Union', 'country', 'Soviet Union')

('amphibious vehicle', 'subclass of', 'land vehicle')

('LuAZ-967', 'country of origin', 'Soviet Union')

('LuAZ-967', 'mass', '950')

Q: Is the wheelbase of the LuAZ-967 equal to 1800?. Make sure to ensure that you do BFS on the graph using $<$think$>$..$<$/think$>$ tags. Enclose your answer in $<$answer$>$...$<$/answer$>$. Answer with only True or False.
\end{mdframed}
\caption{LC-QuAD question, formatted using triples.}
\label{appfig:triple_lcquad}
\end{figure}

\begin{figure}[!h]
\centering
\begin{mdframed}[backgroundcolor=Ivory1]
\scriptsize
In order to determine if the question "Is the wheelbase of the LuAZ-967 equal to 1800?" is true or false, perform a breadth-first search (BFS) on the directed graph. The graph's nodes are labeled with specific words, and the edges represent connections between the nodes. The goal is to find a path through the graph that supports or contradicts the question.

Here is the mapping of node IDs to labels:

\{0: "LuAZ-967", 1: "amphibious vehicle", 2: "Soviet Union", 3: "$<$'wheelbase'\textbar '1800'$>$", 4: "$<$'ride height' \textbar '280'$>$", 5: "$<$'length' \textbar '3682'$>$", 6: "$<$'height' \textbar '1580'$>$", 7: "$<$'subclass of' \textbar 'amphibious vehicle'$>$", 8: "$<$'width' \textbar '1712'$>$", 9: "$<$'country of origin' \textbar 'Soviet Union'$>$", 10: "$<$'mass' \textbar '950'$>$", 11: "$<$'described by source' \textbar 'Armenian Soviet Encyclopedia, vol. 1'$>$", 12: "$<$'subclass of' \textbar 'watercraft'$>$", 13: "$<$'subclass of' \textbar 'land vehicle'$>$", 14: "$<$'country' \textbar 'Soviet Union'$>$"\}

The edges in the graph are: 

0 -$>$ 3, 4, 5, 6, 7, 8, 9, 10

1 -$>$ 11, 12, 13

2 -$>$ 14

Answer with only "True" or "False" based on whether a path exists in the graph which supports the question. Provide your answer in $<$answer$>$..$<$/answer$>$.

Do a BFS on the graph to help answer the question: Is the following statement True or False? Is the wheelbase of the LuAZ-967 equal to 1800?
\end{mdframed}
\caption{LC-QuAD question, formatted to convert triples into a graph structure more like that used during training.}
\label{appfig:idgraph_lcquad}
\end{figure}

\section{Appendix 6: Graph Tokens preliminary experiments}

\begin{table}[!h]
\centering
\begin{tabular}{l c c c c}
Method & Node Count & Node Degree & DFS & BFS \\
\hline
\rowcolor{Ivory2}Random & .100 & .150 & .500 & .500 \\

No CoT & .958 & .364 & .848 & .866 \\
With CoT & .956 & .914 & .986 & .990 \\

\end{tabular}
\caption{Comparison of graph token with/without CoT, 5-20 node graphs}
\label{tab:cot_vs_nocot}
\end{table}

Table \ref{tab:cot_vs_nocot} shows performance differences when we are training the model to output CoT or no CoT, and using CoT shows a clear improvement in performance, especially for the Node Degree task (despite its simplicity). Like prior work, we found that the graph tokens alone struggled to learn the local task of node degree, and even including CoT the performance of this task hovered near 35\% accuracy when trained and tested alone. However, performing multi-task training -- with four different tasks and different types of CoT -- the model was able to also learn to solve the Node Degree task correctly. This suggest an important different compared to prior work such as Perozzi et al. (2024), which indicated limitations of graph tokens with certain encoding strategies to effectively learn some tasks when trained on a single task and without producing a CoT as its output.

\begin{table}[!h]
\centering
\begin{tabular}{l c c c c}
Method & Node Count & Node Degree & DFS & BFS \\
\hline
\rowcolor{Ivory2}Random & .100 & .150 & .500 & .500 \\

2N Tokens & .992 & .466 & .862 & .896 \\
30 Tokens & .956 & .914 & .986 & .990 \\
200 Tokens & .996 & .862 & .984 & .990 \\

\end{tabular}
\caption{Comparison of graph token with/without fixed token size, 5-20 node graphs}
\label{tab:fixed_vs_flex}
\end{table}

Table \ref{tab:fixed_vs_flex} indicates performance when the graph is trained and tested with a fixed number of nodes in the graph versus using a variable number of nodes.
When not using fixed token size, the number of graph tokens is 2N where N is the number of nodes in the graph. Using fixed token size limits the extensibility of the method, but appears to help significantly with tasks like node degree (note that results in Table \ref{tab:cot_vs_nocot} were produced using a fixed token count of 30 as well). 
Using only 30 fixed tokens vs 200 seems to have some effect on performance, but it isn't strictly positive or negative. This indicates that extra unused tokens do not hinder the ability of the graph token approach -- in practice it is likely that the extra tokens simply act in a same manner as tokens learned in the P-tuning approach.

When a fixed number of tokens are being used, the input to the graph encoder (and subsequent output/input to LLM) treats the graph as always having a fixed number of nodes, so different input graphs essentially just are changing the adjacency matrix.

\begin{table}[!h]
\centering
\begin{tabular}{l c c c c}
Method & N. Count & N. Degree & DFS & BFS \\
\hline
\rowcolor{Ivory2}Random & .100 & .150 & .500 & .500 \\

With Labels & .998 & .926 & .976 & .992 \\
Without Labels & .956 & .914 & .986 & .990 \\

\end{tabular}
\caption{Comparison of including node label text in the prompt, 5-20 node graphs}
\label{tab:lab_vs_nolab}
\end{table}

Table \ref{tab:lab_vs_nolab} shows performance comparing when the model is trained/tested while including node labels in the prompt v.s. not including labels. When labels are not included, questions are all phrased to ask about nodes by ID (e.g. "what is the degree of node 7?"), and when including labels we use randomized node names. Including node labels in the prompt appears to somewhat improve performance -- this also is useful to allow us to compare graph tokens against other approaches in a more comparable manner, and also allows for an alternative approach to introduce semantics into the graph nodes without having to devise a complex encoding mechanism. This setting uses significantly more tokens than when no labels associated with the ndoes are in the prompt, but it appears to somehow change what the graph encoder learns to encode and output as graph tokens.

\begin{table}[!h]
\centering
\begin{tabular}{l c c c c}
Method & N. Count & N. Degree & DFS & BFS \\
\hline
\rowcolor{Ivory2}Random & .100 & .150 & .500 & .500 \\

With Labels & .428 & .410 & .720 & .710 \\
Without Labels & .034 & .420 & .764 & .672 \\
\end{tabular}
\caption{Comparison of including node label text in the prompt, on OOD graph sizes containing 20-30 nodes (trained on 5-20 node graphs)}
\label{tab:ood_lab_vs_nolab}
\end{table}

Table \ref{tab:ood_lab_vs_nolab} compares when we include labels in the prompt text vs not, on OOD graph sizes (trained on 5-20 node graphs, tested on 20-30 node graphs). Including labels in the prompt appears to make a significant difference in out of distribution performance for the node count task. As the base training data in this case only trains to output answers for node counts between 5 and 20, poor performance has been seen quite often. However, what the graph encoder learns to encode when labels are in the prompt somehow affects how well it generalizes to output numbers that were never seen during training. 

Given the results of these explorations, the main experiments using graph tokens were conducted using a fixed node count, and node labels were included in the prompt.

\section{Appendix 7: OOD Length Experiments}

The following tables show results for test data where graphs has no limitation on the number of edges and the answers to the test questions tended to require longer CoTs than questions seen during training. We generated test data to ensure that we have an even distribution of test cases based on the length of the ground-truth CoT, the length of the BFS or DFS trace lengths, or the length of the shortest path connecting nodes (for reachability). The distribution of these elements in this test data differs from the distribution of data that naturally was produced when generating our training data.

\begin{table}[!h]
\centering
\begin{tabular}{l c c}
Method & DFS & BFS \\
\hline
LoRA & .992 & .996 \\
P-Tuning & .830 & .770 \\
Graph Tokens & .870 & .874 \\
Graph Tok + Text & .778 & .723 \\
\end{tabular}
\caption{Tests on graphs with longer DFS and BFS trace lengths.}
\label{tab:mainres_trace_spread}
\end{table}

\begin{table}[!h]
\centering
\begin{tabular}{l c c}
Method & DFS & BFS \\
\hline
LoRA & .994 & 1.00 \\
P-Tuning & .954 & .958 \\
Graph Tokens & .946 & .944 \\
Graph Tok + Text & .946 & .940 \\
\end{tabular}
\caption{Tests on graphs with longer CoT answer lengths.}
\label{tab:mainres_anslen_spread}
\end{table}

\begin{table}[!h]
\centering
\begin{tabular}{l c c}
Method & DFS & BFS \\
\hline
LoRA & .993 & .998 \\
P-Tuning & .967 & .956 \\
Graph Tokens & .955 & .955 \\
Graph Tok + Text & .957 & .944 \\
\end{tabular}
\caption{Tests on graphs with longer shortest-path lengths (for reachable nodes).}
\label{tab:mainres_pathlen_spread}
\end{table}

In general, we continue to observe very strong performance by LoRA, while other models begin to show relatively weaker performance across these tests. The increased number of edges which are present in these graphs likely contributes to some of the difficulty, as training is only conducted on graphs with up to 500 edges. The other point of difficulty is that of producing longer answers, demonstrating that models besides LoRA do not generalize as well to such test cases. This is most apparent in Table \ref{tab:mainres_trace_spread}, where no model besides LoRA achieves over 90\% accuracy. In the training graphs, we observed that over 75\% of test cases had trace lengths of 19 nodes or fewer, and therefore many of the test cases here are ``out of distribution'' compared to the typical traces which the models learned to solve during training.

It is worth noting that trace length, CoT answer length, and shortest path length are not necessarily equivalent. For example even if two nodes are very close in the graph, searching for a path using DFS or BFS may lead to exploring many nodes before finding the shortest path.

\section{Appendix 8: Model Transfer Across LLMs}

While LoRA convincingly showed the best performance across tasks and in generalizability, we are also interested in how an architecture like Graph Tokens might have benefits in that it is detached from the LLM architecture. Consider, for example, that existing work which combines image encoders with LLMs often start from large, pretrained image encoding models. The benefit of such an approach is that an encoding model trained separately can be plugged in together with an LLM. In our case, we did not have such a pretrained graph encoder, but the final result of our training is a graph encoding model which is suitable for making an LLM better at solving graph tasks (albeit with some caveats on its OOD performance). 

Our exploration, then aims to investigate whether our Graph Token model can be ``portable'' across different LLMs -- i.e., train our Graph Token model on one LLM and apply it to another. If such a transfer of a learned Graph Token model can be successful, it would provide more merit to the approach: LoRA may still be the most effective model when applied to fully train a particular LLM, but Graph Tokens might still provide benefits in terms of training time or scaling. 

We conduct a small set of experiments on two ideas. The first is to investigate whether we can train a Graph Token model on a small LLM and transfer it to a larger LLM, where both LLMs share the same model architecture. Here, we use the Granite 3.3 family of LLMs for our experiments, training our Graph Token model using the 2B\footnote{https://huggingface.co/ibm-granite/granite-3.3-2b-instruct} model and then trying to transfer it to the 8B\footnote{https://huggingface.co/ibm-granite/granite-3.3-8b-instruct} model. The experimental procedure is to first train our Graph Token on the Granite3.3 2B model, similarly to our other experiments. Then we would freeze the weights of our graph encoder, and train only the final projection layer togher with Granite3.3 8B. Our intuition here is that by training on a smaller model, we can cut down significantly on computational cost of training our Graph Token's encoder, and then continuing training on only the final projection layer can be completed with much less training. 

\begin{table}[!h]
    \centering
    \begin{tabular}{l l l l l}
    Model & Frozen Enc.? & Epochs & DFS & BFS \\
    \hline
    Granite3.3 2B & \xmark & 40 & .996 & .980 \\
    Granite3.3 8B & \xmark & 40 & .986 & .966 \\
    Granite3.3 8B & \cmark & 1 & .966 & .998 \\
    \end{tabular}
    \caption{Transfer experiment of trained Graph Token model, trained on a 2 billion parameter model and transferred to a 8 billion parameter model.}
    \label{apptab:2bto8b}
\end{table}

Table \ref{apptab:2bto8b} shows experimental results, where we compare model performance when we train a Graph Token model from scratch on Granite3.3 2B and Granite3.3 8B, indicated by ``Frozen Encoder? = \xmark''. Both models reach nearly identical performance. We then compare the performance of transferred our Graph Token model -- trained for 40 epochs on Granite 2B -- to the Granite 8B model, leaving the graph encoder weights frozen and training only the projection layer for 1 epoch. We can see that this simple transfer approach enables the Graph Token model to be applied to Granite 8B, achieving nearly identical performance without the need for extensive training. Even beyond just the difference in training epochs, Granite 8B also requires significantly more computational resources to train than Granite 2B, which makes the potential benefits of such a model transfer even better.

Our second experiment aims to investigate whether we can train a Graph Token model on one LLM and transfer to another LLM, where the two LLMs have different model architectures. This time, we perform our transfer experiment by training a Graph Token model using Granite3.3 2B and seeing if it can be transferred to the Phi4-Mini-Instruct model, which we use throughout our main experiments. These results can be seen in Table \ref{apptab:granitetophi}. We see that here, the transfer is not quite as successful as with the transfer among LLMs of the same model family. We do still observe that Phi4 can achieve 95\% accuracy on the BFS task, but the DFS performance is clearly weaker. Transfer among different model architectures remains an interesting direction for future work.

\begin{table}[!h]
    \centering
    \begin{tabular}{l l l l l}
    Model & Frozen Enc.? & Epochs & DFS & BFS \\
    \hline
    Granite3.3 2B & \xmark & 40 & .968 & .986 \\
    Phi4 Mini & \xmark & 40 & .988 & .990 \\
    Phi4 Mini & \cmark & 1 & .736 & .958 \\
    \end{tabular}
    \caption{Transfer experiment of trained Graph Token model, trained on a 2 billion parameter Granite model and transferred to a 4 billion parameter Phi model.}
    \label{apptab:granitetophi}
\end{table}

\end{document}